\documentclass[11pt]{article}
\pdfoutput=1

\usepackage{authblk}
\usepackage[final]{acl}
\usepackage{relsize}
\usepackage{times}
\usepackage{latexsym}
\usepackage{tabularx}
\usepackage{booktabs} 
\usepackage[T1]{fontenc}

\usepackage[utf8]{inputenc}

\usepackage{microtype}

\usepackage{inconsolata}
\usepackage{amsmath}
\usepackage{amsfonts}
\usepackage{footnote}
\usepackage{booktabs}
\usepackage{hyperref} 
\usepackage{multirow}
\usepackage{colortbl}
\usepackage{xcolor}
\usepackage{subfigure}

\usepackage{graphicx}
\usepackage{makecell}
\usepackage[final]{acl}

\usepackage[T1]{fontenc}
\usepackage[utf8]{inputenc}
\usepackage{microtype}
\usepackage{inconsolata}
\usepackage{booktabs}
\usepackage{amsmath}
\usepackage{amsfonts}
\usepackage{xcolor}
\definecolor{chessgreen}{HTML}{769656}
\newcommand{\greenterm}[1]{\textcolor{chessgreen}{\textbf{#1}}}
\usepackage{graphicx}
\usepackage{placeins}

\title{Do Chess Explanations Reflect Model Decisions?\\
Behavioral and Token-Level Tests of LLM Reasoning Faithfulness}

\author[1,2]{\textbf{Angelina Parfenova}}

\affil[1]{Lucerne University of Applied Sciences and Arts}
\affil[2]{Technical University of Munich}

\begin{document}

\maketitle
\begin{abstract}
Large language models can produce fluent explanations for chess moves, but plausible language does not necessarily reflect the reasoning behind a decision.
We study this question in chess, where the board state is fully observable, legal actions can be enumerated, and move quality can be evaluated independently.
Across 200 Lichess endgame puzzles, we test explanations using move recoverability, decoder-side controls, and token-level scoring of legal candidate moves.
Unmasked explanations make generated moves easy to recover, but this advantage drops sharply after explicit move hints are removed.
Under strict masking, explanations provide only small and decoder-dependent gains over the board state alone.
Token-level scoring shows that explanations can nevertheless alter move preferences: random but plausible explanations from other puzzles reduce the probability of the correct move, indicating that irrelevant reasoning text is not simply ignored.
We also find that recognizable endgame motifs can make generated moves easier to recover without reliably improving move correctness.
Together, these results show that linguistic plausibility, consistency with a generated action, and solution correctness are distinct properties.
Fluent chess explanations can influence action preferences and support a coherent move narrative while providing only limited evidence of faithful reasoning.
\end{abstract}

\section{Introduction}

LLMs are increasingly asked not only to answer questions, but also to explain their answers, especially through step-by-step rationales and chain-of-thought-style prompting \citep{wei2022chain, wang2022self}.
For many tasks, however, explanations are difficult to evaluate: multiple rationales may sound plausible, and the underlying decision process is not directly observable.
Prior work has shown that explanations can be persuasive without being faithful to the model's actual decision process \citep{jain2019attention, jacovi2020towards}, and that models may rationalize answers after the fact rather than disclose the features that shaped their prediction \citep{turpin2023language, lanham2023measuring}.
This creates a central interpretability problem: is a model's explanation consistent with the preference that produced its answer, or is it a fluent post-hoc narrative?

\begin{figure}
\centering
\includegraphics[width=\linewidth]{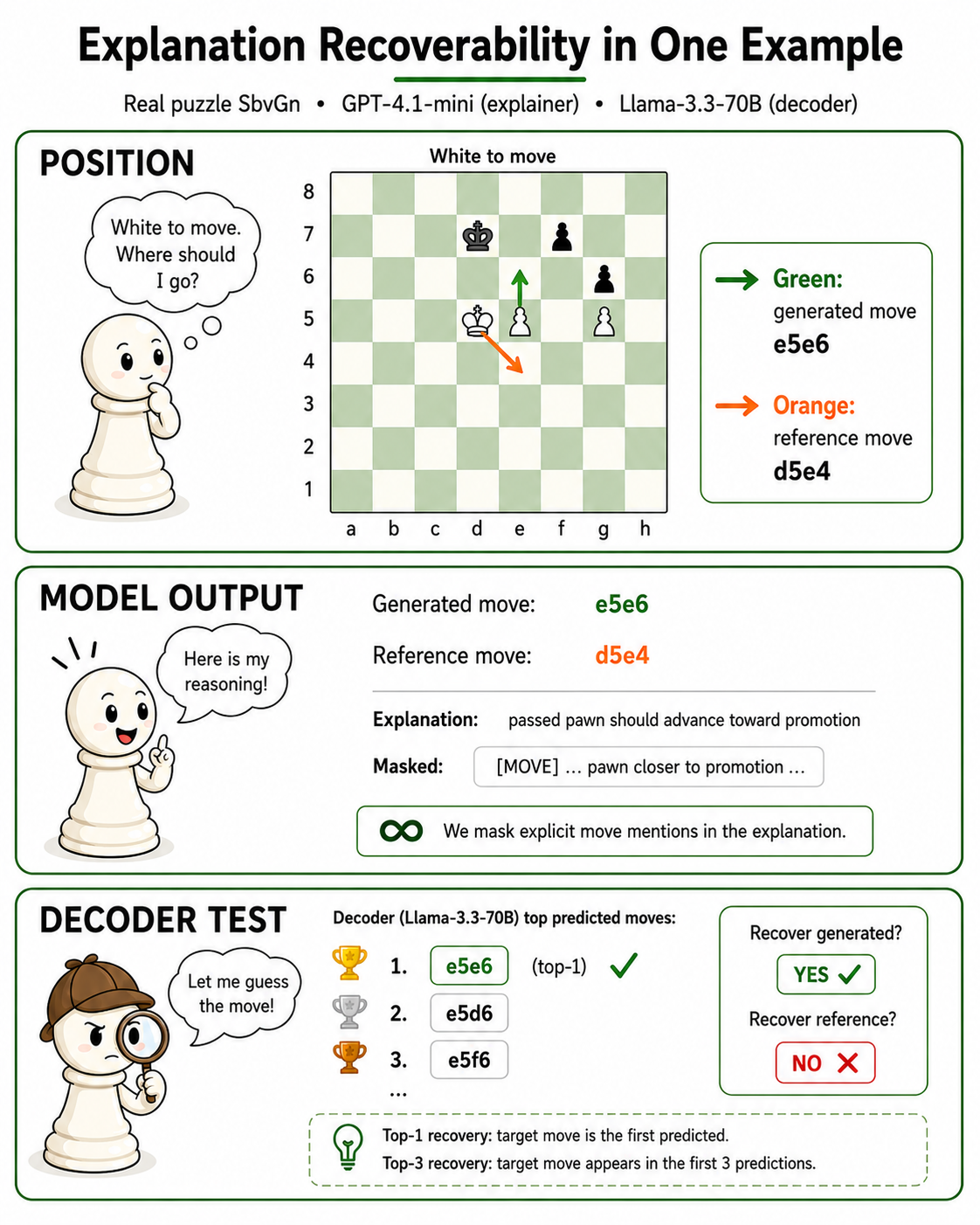}
\caption{
Example of explanation recoverability.
After masking explicit move mentions, the decoder recovers the model's generated move (\texttt{e5e6}) but not the reference move (\texttt{d5e4}), showing action-specific information without solution correctness.
}
\label{fig:recoverability_example}
\end{figure}

Chess provides a useful controlled setting for this question.
It has long been used to study human expertise, search, and pattern recognition \citep{de2008thought, chase1973perception}, and recent work has used chess to evaluate LLM board-state comprehension, move legality, and strategic reasoning \citep{kuo2023large, feng2023chessgpt, liu2026chessarena}.
A chess position defines a concrete state, legal actions are enumerable, move quality can be checked with engines, and human experts routinely explain moves through tactical lines, motifs, and opponent responses.
Unlike open-ended domains, chess lets us ask whether an explanation contains enough state-specific information to identify the move it claims to justify.

We study the relation between explanation-action consistency and token-level action preference.
Our central behavioral metric is \textit{move recoverability}: given a chess position and a masked explanation with explicit move mentions removed, can a decoder infer the intended move?
If an explanation faithfully communicates the decision-relevant structure of a move, the move should be recoverable from that explanation.
Recoverability is therefore a behavioral proxy for faithfulness, not direct evidence about the model's internal causal computation.
We then ask whether recoverability tracks token-level preference: does an open scorer rank the reference move highly, and does adding an explanation make either the generated or reference move more probable?
This token-level view complements behavioral recovery while avoiding the stronger claim that free-text explanations directly reveal internal causal reasoning \citep{belrose2023eliciting}.

We make five contributions:
(1) we introduce move recoverability as a behavioral interpretability test for chess explanations;
(2) we measure token-level move preference by scoring legal candidate moves under open-weight models;
(3) we separate recovery and scoring of the generated move from the reference move;
(4) we introduce FEN-only, explanation-only, random-explanation, masked, and unmasked controls for both decoder recovery and all-legal token-level scoring;
and (5) we analyze endgame-specific motif language and compare LLM explanations with controlled and naturalistic human chess reasoning transcripts.

\section{Related Work}

\textbf{Chess expertise and early LLM benchmarks.}
Chess has long been used to study human expertise, search, and pattern recognition \citep{de2008thought,chase1973perception}.
Endgame studies provide curated positions with known solutions \citep{livshitz1988}, and modern engines provide strong move evaluation \citep{stockfish}.
Early LLM studies evaluated board-state comprehension, legality, and move quality \citep{kuo2023large}, while ChessGPT combined game trajectories with chess-language corpora for policy and language learning \citep{feng2023chessgpt}.
These works establish chess as a structured reasoning benchmark, but do not evaluate whether free-text explanations faithfully identify the actions they justify.

\textbf{Language-augmented chess reasoning.}
MATE provides roughly one million Lichess positions whose candidate moves are annotated with strategic and tactical information, and uses these annotations to supervise move selection \citep{wang2025explore}.
Concept-guided Chess Commentary combines expert-model concepts with an LLM to produce fluent move commentary and evaluates its informativeness and linguistic quality \citep{kim2025bridging}.
Both lines demonstrate the value of chess language for supervision and pedagogy, but their explanations are generally unmasked and conditioned on visible moves or candidates.

\textbf{Visible and agentic chess traces.}
Recent chess benchmarks evaluate models through games, move selection, puzzle solving, or interactive tool use \citep{liu2026chessarena,kolasani2025llm}.
Other work distills move annotations or square-grounded explanations into chess-specialized models \citep{tang2026grounded}.
These studies test whether traces improve task performance or instruction following, whereas we test whether explanation text still identifies the intended action after move-specific leakage is masked.
Because aligned human and LLM think-aloud corpora on the same positions remain scarce, our human comparison is exploratory rather than a matched performance benchmark.

\textbf{Faithfulness of explanations.}
Prior work shows that plausible explanations need not be faithful to the model's actual decision process \citep{jain2019attention,jacovi2020towards}.
Although step-by-step prompting can improve performance \citep{wei2022chain}, models may still rationalize biased or perturbed answers after the fact \citep{turpin2023language,lanham2023measuring}.
These findings motivate behavioral tests of explanation faithfulness.
We contribute a functional test in chess: whether a masked explanation permits recovery of the action it explains and shifts token-level preference toward that action.

\textbf{Reasoning stability and representation probes.}
Self-consistency improves answer accuracy by aggregating across sampled reasoning paths, but agreement among outputs does not by itself establish that any individual rationale is faithful \citep{wang2022self}.
Mechanistic probes offer a complementary view by decoding intermediate representations.
However, direct logit-lens projections can be sensitive to representational basis, and tuned probes can differ substantially from the raw output projection \citep{belrose2023eliciting}.
We therefore treat token scoring and our layerwise analysis as descriptive evidence about local model preference, not as a causal reconstruction of reasoning.

\section{Experimental Setting}

\subsection{Puzzle Data}

We evaluate primarily on a 200-puzzle subset of the Lichess puzzle database \citep{lichessDB}.
Puzzles are filtered to endgames, restricted to ratings 500-2000, and labeled into human-interpretable endgame categories, such as pawn, rook, queen, bishop, and knight endings, using Lichess themes and material heuristics. Each puzzle provides a FEN position and a reference solution move.
We also use a six-puzzle pawn-endgame pilot from \textit{Test Your Endgame Ability} \citep{livshitz1988} for controlled human comparison.

\subsection{LLM Explanations}

For each puzzle, we prompt models under three styles: brief explanation, calculation-style reasoning, and teaching explanation.
The hosted-model study covers six model families-GPT-4.1-mini, Llama-3.1-8B-Instant, Llama-3.3-70B-Versatile, Llama-4-Scout-17B, Kimi-K2-Instruct, and Qwen3-32B-for 18 model-prompt conditions.
We additionally run token-level analyses with Mistral-7B, Qwen2.5-32B, and Llama-3.1-70B.
For each generation we record the parsed move, legality, exact match to the reference move, Stockfish evaluation difference, and explanation text.
Appendix Tables~\ref{tab:hosted_model_summary} and \ref{tab:engine_clean} report the hosted-model summary and cleaned GPT-4.1-mini move-quality analysis.



\section{Move Recoverability}

Move recoverability evaluates whether an explanation communicates an identifiable action.
Given a FEN position and an explanation, we mask explicit move strings, SAN/UCI mentions, and direct lexical leakage under a strict masking rule.
We then ask one of three decoder models to infer the intended move as a UCI string: GPT-4.1-mini, Llama-3.3-70B, or Qwen3-32B.
We report top-1 and top-3 recoverability for both generated-move and reference-move targets.

\begin{figure}[h]
    \centering
    \includegraphics[width=0.85\columnwidth]{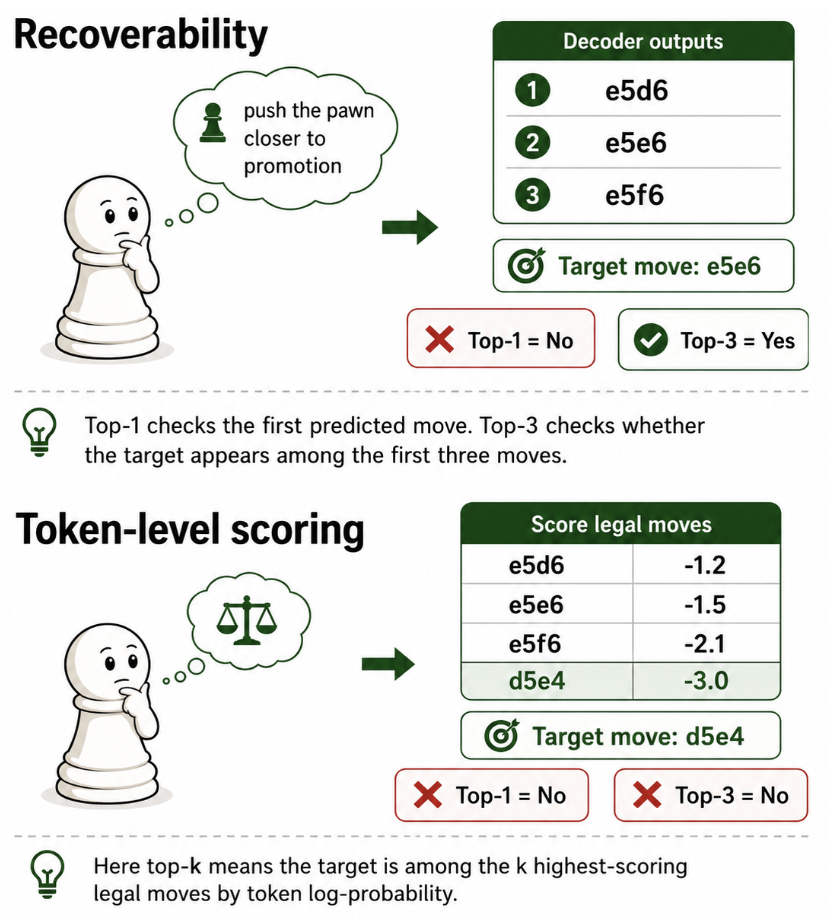}
    \caption{
    Illustration of top-$k$ evaluation in our experiments.
    In recoverability, a decoder predicts ranked move candidates from a masked explanation, and top-$k$ checks whether the target move appears among the first $k$ outputs.
    In token-level scoring, legal moves are ranked by summed token log-probability, and top-$k$ checks whether the target move appears among the $k$ highest-scoring candidates.
    }
    \label{fig:topk_explanation}
\end{figure}

\[
\mathrm{Recoverability} =
\frac{\# \text{correctly inferred moves}}{\# \text{explanations}}.
\]

We distinguish two targets.
\greenterm{Generated-move recoverability} measures whether the decoder reconstructs the move produced by the explanation-generating model.
\greenterm{Reference-move recoverability} measures whether the decoder reconstructs the Lichess solution. Table~\ref{tab:recoverability_targets} shows that these are substantially different.
Generated-move top-1 recoverability is 0.379-0.398, whereas reference-move top-1 recoverability is only 0.056-0.091.
Sample sizes vary slightly because rows without a parseable generated move or usable explanation are excluded for the corresponding target.
The calculation prompt produces much longer explanations and higher specificity scores, but does not improve recoverability.
Thus, explanations communicate the model's selected action considerably better than they communicate the correct action.

\begin{table}[t]
\centering
\scriptsize
\setlength{\tabcolsep}{3pt}
\renewcommand{\arraystretch}{0.95}
\begin{tabular}{@{}llcc@{}}
\toprule
Prompt & Target & Top-1 & Top-3 \\
\midrule
Brief ($n=191$) & Gen. & 0.398 [.330,.466] & 0.518 [.450,.592] \\
                & Ref. & 0.058 [.026,.094] & 0.099 [.058,.147] \\
\addlinespace[1pt]
Calc. ($n=195$) & Gen. & 0.379 [.313,.446] & 0.462 [.390,.533] \\
                & Ref. & 0.056 [.026,.092] & 0.103 [.062,.149] \\
\addlinespace[1pt]
Teach. ($n=197$) & Gen. & 0.391 [.325,.462] & 0.503 [.431,.574] \\
                 & Ref. & 0.091 [.056,.132] & 0.137 [.091,.188] \\
\bottomrule
\end{tabular}
\caption{
Strict-mask recoverability with puzzle-bootstrap 95\% intervals.
Gen. targets the GPT-4.1-mini generated move; Ref. targets the Lichess solution.
}
\label{tab:recoverability_targets}
\end{table}

Cross-model reruns with Llama-3.3-70B and Qwen3-32B preserve the generated-over-reference pattern for every prompt (Appendix Tables~\ref{tab:decoder_recoverability_200}-\ref{tab:decoder_recoverability_200_parsed}).
However, absolute generated-move top-1 recoverability drops from roughly 0.38-0.40 with GPT-4.1-mini to 0.067-0.077 with Llama and 0.057-0.097 with Qwen.
Thus, changing the decoder greatly reduces absolute recovery rates, but generated moves remain more recoverable than reference moves.

\subsection{Decoder Controls and Robustness}

To separate explanation signal from decoder chess ability, we run six conditions on the same 50-puzzle subset used for all-legal scoring: FEN only, explanation only, FEN plus strict masking, FEN plus maximal masking, FEN plus a random same-style explanation, and FEN plus the unmasked explanation.
We evaluate Llama-3.3-70B and Qwen3-32B as cross-model decoders.
Decoder outputs are constrained to UCI move predictions and unparsed outputs are counted as failures.

Table~\ref{tab:decoder_controls_groq} shows a consistent pattern across the two decoders.
Masked explanations provide little improvement over the FEN-only baseline: generated-move recovery increases only slightly, and the paired bootstrap intervals include zero.
Reference-move recovery remains low in all masked conditions.
In contrast, unmasked explanations make the generated move much easier to recover, showing that explicit move notation is a major source of recovery when left in the text.
Explanation-only and random-explanation controls perform near zero, suggesting that recovery is not driven by generic chess language alone.
Overall, these controls suggest that masked explanations contain at most a modest signal beyond the board state, while unmasked explanations mostly reveal the answer directly.
\begin{table*}[t]
\centering
\scriptsize
\begin{tabular}{llccccc}
\toprule
Decoder & Condition & Parsed & Generated@1 & Generated@3 & Reference@1 & Reference@3 \\
\midrule
Llama-3.3-70B & FEN only & 0.510 & 0.047 & 0.060 & 0.000 & 0.000 \\
 & Explanation only & 0.201 & 0.000 & 0.020 & 0.000 & 0.007 \\
 & Strict mask & 0.436 & 0.073 & 0.100 & 0.053 & 0.053 \\
 & Maximal mask & 0.423 & 0.073 & 0.073 & 0.060 & 0.060 \\
 & Random expl. & 0.376 & 0.007 & 0.013 & 0.000 & 0.007 \\
 & Unmasked & 0.879 & 0.680 & 0.740 & 0.113 & 0.147 \\
\midrule
Qwen3-32B & FEN only & 0.376 & 0.007 & 0.007 & 0.007 & 0.020 \\
 & Explanation only & 0.121 & 0.000 & 0.007 & 0.007 & 0.007 \\
 & Strict mask & 0.369 & 0.020 & 0.033 & 0.020 & 0.027 \\
 & Maximal mask & 0.450 & 0.027 & 0.027 & 0.027 & 0.027 \\
 & Random expl. & 0.389 & 0.013 & 0.020 & 0.013 & 0.013 \\
 & Unmasked & 0.826 & 0.737 & 0.743 & 0.113 & 0.113 \\
\bottomrule
\end{tabular}
\caption{Decoder-side controls on the same 50-puzzle subset, pooled across prompt styles. Parsed is the rate at which the decoder returned at least one extractable move. Qwen3 uses hidden/no-reasoning mode.}
\label{tab:decoder_controls_groq}
\end{table*}


\section{What Signal Do Explanations Carry?}

The recoverability results show that masked explanations preserve more information about the generated move than about the reference solution.
We now ask what kind of signal this is.
We test whether explanations shift token-level candidate rankings, whether recoverability tracks token-level preference, whether familiar endgame motifs explain the effect, and whether reference-move salience increases across model layers.

\subsection{Token-Level Candidate Scoring}

Recoverability measures whether a decoder can infer a move from explanation text.
This does not show whether the underlying language model assigns high probability to the reference move.
We therefore separately score candidate moves by conditional token log-probability under open-weight models. We score candidate moves by conditional log-probability under a minimal prompt:
\begin{quote}
\small
Position (FEN): \texttt{<fen>}\\
Best move in UCI:
\end{quote}
Candidate sets include the reference move, generated move when available, Stockfish move, and legal distractors.
Because chess moves may tokenize into multiple pieces, we sum token log-probabilities over the full UCI string.
The 200-puzzle analyses use filtered candidate sets.
We additionally score \emph{every legal move} on a 50-puzzle subset for the controlled explanation experiment below.

Across open scoring models, reference moves are not strongly preferred.
Under scoring-only prompts, Qwen2.5-32B ranks the reference move first in 0.245 of puzzles and in the top three in 0.555, while Llama-3.1-70B reaches 0.285 top-1 and 0.575 top-3.
The smaller style-conditioned scorer shows the same pattern: reference top-1 ranges from 0.175 to 0.215 across prompt styles, while top-3 ranges from 0.49 to 0.54.
Thus, open models often assign some probability mass to the reference move, but rarely rank it decisively first.
Style-conditioned scoring changes these ranks only modestly, suggesting that prompt style affects surface explanation more than token-level move preference.
Detailed rank statistics are reported in Appendix~A.2.



\subsection{Controlled Explanation-Conditioned Scoring}

If an explanation communicates decision-relevant information, conditioning another model on it should make the explained move easier to rank.
We therefore use Qwen2.5-32B as a cross-model scorer for all legal moves on 50 puzzles.
The explanations and generated moves come from GPT-4.1-mini.
We compare four contexts: FEN only, FEN plus a maximally masked explanation, FEN plus the unmasked explanation as a leakage upper bound, and FEN plus a random explanation from another puzzle under the same prompt style.
Maximal masking removes UCI/SAN, coordinates, piece-square references, promotion terms, and direct move phrases.
\begin{quote}
\small
Position (FEN): \texttt{<fen>}\\
Explanation: \texttt{<masked explanation>}\\
Best move in UCI:
\end{quote}

Table~\ref{tab:qwen32b_controls} and Appendix Figure~\ref{fig:qwen32b_controls} show that maximal masking does not reliably improve reference ranking.
Relative to FEN only, reference top-3 changes by $-0.06$, $+0.02$, and $0.00$ for brief, calculation, and teaching prompts; all paired 95\% intervals include zero.
Generated-move top-3 changes by $+0.08$, $+0.04$, and $+0.12$.

The controls clarify the interpretation.
Unmasked explanations increase generated-move top-1 by $+0.42$, $+0.30$, and $+0.50$, confirming direct lexical leakage, but do not improve reference ranking.
Random explanations reduce reference top-3 for brief ($-0.16$, $[-0.30,-0.04]$) and calculation ($-0.18$, $[-0.30,-0.06]$).
Thus, explanation text can steer token-level preference, but its reliable signal is more closely tied to the generated action than to solution correctness.
\begin{table}[t]
\centering
\scriptsize
\setlength{\tabcolsep}{2.5pt}
\begin{tabular}{llcccc}
\toprule
Prompt & Context & R@1 & R@3 & G@1 & G@3 \\
\midrule
Brief & FEN only & 0.16 & 0.42 & 0.08 & 0.22 \\
 & + maximal mask & 0.14 & 0.36 & 0.12 & 0.30 \\
 & + unmasked & 0.18 & 0.36 & 0.50 & 0.52 \\
 & + random expl. & 0.06 & 0.26 & 0.06 & 0.20 \\
\midrule
Calculation & FEN only & 0.18 & 0.40 & 0.10 & 0.16 \\
 & + maximal mask & 0.18 & 0.42 & 0.10 & 0.20 \\
 & + unmasked & 0.12 & 0.38 & 0.40 & 0.42 \\
 & + random expl. & 0.04 & 0.22 & 0.04 & 0.16 \\
\midrule
Teaching & FEN only & 0.16 & 0.36 & 0.10 & 0.22 \\
 & + maximal mask & 0.16 & 0.36 & 0.14 & 0.34 \\
 & + unmasked & 0.14 & 0.36 & 0.60 & 0.60 \\
 & + random expl. & 0.14 & 0.30 & 0.08 & 0.26 \\
\bottomrule
\end{tabular}
\caption{Qwen2.5-32B all-legal scoring on 50 puzzles. R and G denote reference and generated moves. Unmasked explanations are a leakage check; random explanations come from another puzzle under the same prompt style.}
\label{tab:qwen32b_controls}
\end{table}

\begin{figure*}[t]
    \centering
    \includegraphics[width=0.8\textwidth]{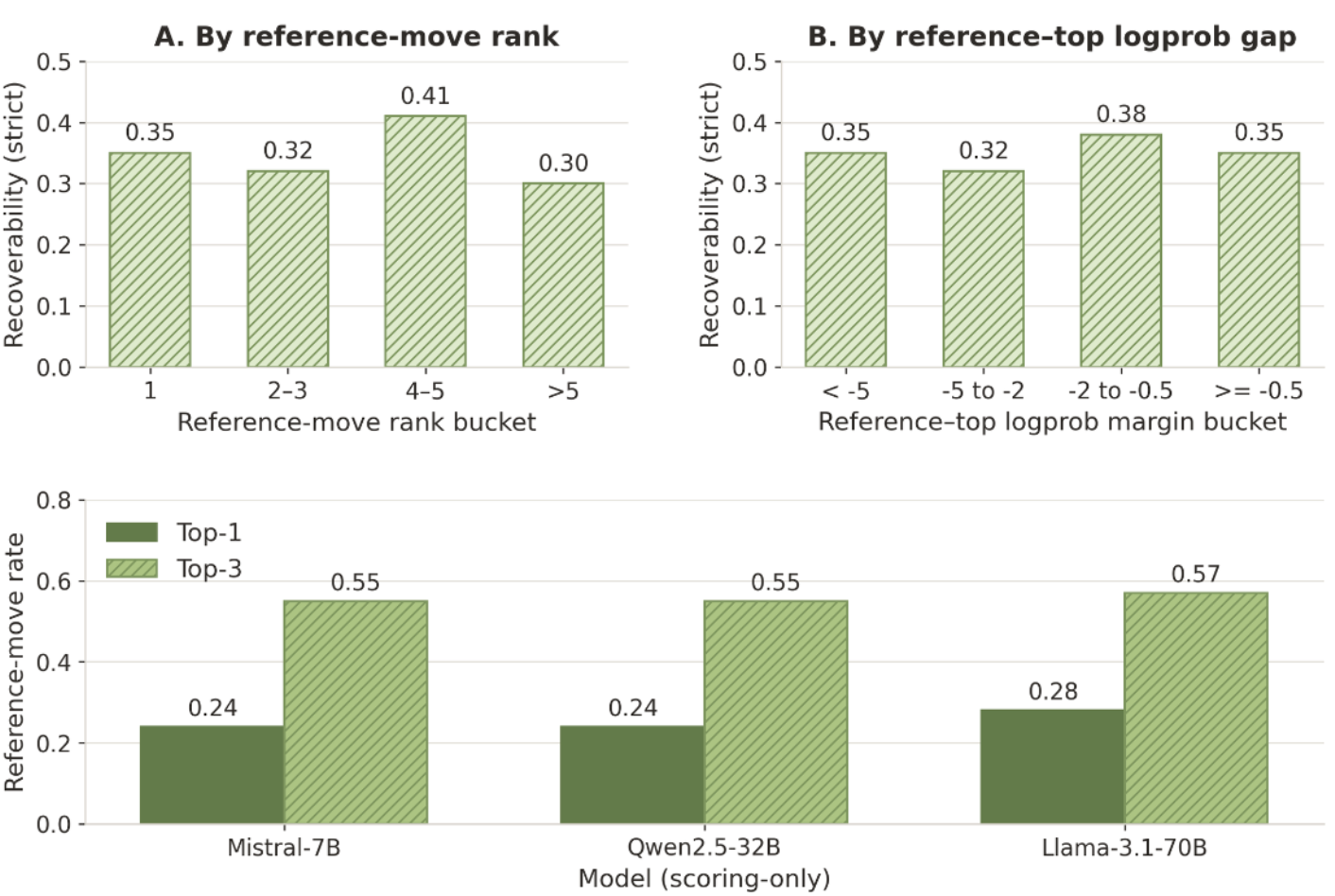}
    \caption{
    Token-level preference and recoverability.
    (A) Recoverability by reference-move rank bucket.
    (B) Recoverability by reference-versus-top log-probability gap.
    Both relationships are weak and non-monotonic, showing that recoverable explanations do not necessarily correspond to stronger token-level preference for the reference move.
    (C) Reference-move top-1 and top-3 rates across open scoring models under scoring-only prompts.
    Open models place the reference move in the top three more often than top one, but rarely rank it decisively first.
    }
    \label{fig:token_preference_recoverability}
\end{figure*}

\begin{table*}[t]
\centering
\small
\begin{tabular}{lrrrrrr}
\toprule
Endgame type & Puzzles & Legal & Exact & Gen.\ rec. & Ref.\ rec. & Motif mention \\
\midrule
Pawn & 40 & 0.558 & 0.133 & 0.483 & 0.125 & 0.775 \\
Rook & 56 & 0.467 & 0.103 & 0.494 & 0.062 & 0.788 \\
Knight & 10 & 0.700 & 0.067 & 0.233 & 0.033 & 0.867 \\
Bishop & 14 & 0.300 & 0.025 & 0.256 & 0.026 & 0.600 \\
Queen$^\dagger$ & 7 & 0.810 & 0.190 & 0.238 & 0.143 & 1.000 \\
Queen-rook & 65 & 0.558 & 0.037 & 0.323 & 0.053 & 0.979 \\
Bishop-knight$^\dagger$ & 7 & 0.211 & 0.000 & 0.263 & 0.000 & 0.579 \\
Bishop vs.\ pawns$^\dagger$ & 1 & 1.000 & 0.667 & 0.333 & 0.000 & 1.000 \\
\bottomrule
\end{tabular}
\caption{Endgame-type analysis over the three GPT-4.1-mini prompt conditions. Recoverability is top-1. $^\dagger$Fewer than ten puzzles; descriptive only.}
\label{tab:endgame_summary}
\end{table*}

\subsection{Linking Preference and Recoverability}

We join token-level rank data with recoverability outcomes to test whether explanations are more recoverable when the reference move receives a higher token-level rank.
Figure~\ref{fig:token_preference_recoverability} shows that recoverability is only weakly related to token-level preference, while open scoring models rarely rank the reference move first.
Detailed bucket counts and engine-loss summaries are reported in Appendix~\ref{sec:preference_recoverability_details}.

This is a central result: recoverable explanations do not necessarily correspond to stronger token-level preference for the reference move.
They contain partial decision-relevant information, but not enough to reliably reconstruct the move, and not in a way that strongly tracks the model's candidate ranking.

\section{Endgame Types and Motif Language}

We next test whether familiar verbal schemas are associated with stronger reasoning.
Using material and Lichess-theme labels, we group positions into pawn, rook, knight, bishop, queen, queen-rook, bishop-knight, and bishop-versus-pawn endings.
We treat motif vocabularies as overlapping lexical schemas: tactical (forks, checks, captures, pins), strategic (passed pawns, promotion races, simplification), positional (king or piece activity, blockade, file and color-complex control), and endgame-theoretical (opposition, triangulation, zugzwang, Lucena, Philidor, and the square of the pawn).
These non-exclusive categories distinguish broad verbal schemas from exact move calculation.
Configurable motif vocabularies cover promotion and passed pawns, opposition and king activity, knight forks and tempi, rook activity and checking, bishop diagonals and color complexes, and queen perpetual-check language.
For each explanation, we record expected motif mentions, cross-category motif mentions, and missing expected motifs.

Table~\ref{tab:endgame_summary} shows substantial differences.
We restrict comparisons to categories with at least ten puzzles.
Among the large categories, generated-move recoverability is highest in rook (0.494) and pawn endings (0.483), where conventional verbal schemas are well established, but exact reference accuracy remains only 0.103 and 0.133.
Queen-rook endings are frequent but weaker on both exact accuracy (0.037) and generated recovery (0.323).
Queen endings show a high exact rate in this sample, but the category is too small for interpretation.
Reference recovery is far below generated recovery in every sufficiently large category.


Motif mention is not equivalent to correct reasoning.
In pawn endings, mentioning an expected motif is associated with generated-move recoverability increasing from 0.259 to 0.548, while exact accuracy changes only from 0.111 to 0.140.
In rook endings, motif mention similarly raises generated recovery from 0.400 to 0.520 but exact accuracy decreases from 0.143 to 0.092.
The exploratory multivariable classifier likewise gives expected-motif mention a small generated-recovery coefficient whose bootstrap interval includes zero (Table~\ref{tab:predictive_selected}).
The motif results therefore suggest that LLMs often activate recognizable tactical, strategic, positional, or endgame-theoretical schemas, but that this schema activation does not reliably translate into correct move selection.

\section{Layerwise Exploratory Analysis}

As a secondary representational check, we ask whether the reference move becomes more salient across layers in open-weight models.
For 20 puzzles, we apply the model output head to each layer's hidden state and compute the log-probability of the complete reference-move UCI sequence.
This analysis is not meant to reconstruct the model's causal reasoning process.
Instead, it asks a narrower question: does the correct move become increasingly compatible with the model's intermediate representations as computation proceeds?

Both tested models show the same qualitative pattern.
For Llama-3.1-70B, the mean reference-move log-probability rises from $-47.59$ at the first layer to $-21.46$ at the final layer.
For Qwen2.5-32B, it rises from $-48.26$ to $-17.64$.
Thus, the reference move becomes less unlikely in later layers, suggesting that the model representations increasingly encode some information compatible with the correct action.
However, the final-layer probability remains weak in absolute terms, and the reference move is often not the top-ranked candidate.
The increase should therefore not be read as evidence that the model has selected the correct solution.

This result complements the behavioral findings.
Move recoverability shows that explanations can contain partial information about a generated action, while token-level scoring shows that the correct reference move is often not decisively preferred.
The layerwise analysis adds a representational view: correct-move salience can increase during the forward pass without becoming strong enough to dominate the model's final move preference.

\section{Human and LLM Grounding Styles}

We conduct an exploratory comparison of how human and LLM explanations refer to the chessboard.
This analysis is not intended as a human-model performance comparison: the sources differ in positions, recording conditions, explanation length, and segmentation.
Instead, we ask whether human and model explanations expose similar kinds of board-grounded information, such as explicit coordinates, candidate moves, spatial references, and chess concepts.

We extract four segment-level marker types.
\textit{Coordinate} markers include explicit board squares or ranks and files, such as ``e5'', ``d4'', ``seventh rank'', or ``g-file''.
\textit{Move} markers include move notation or move-like action descriptions, such as ``e5e6'', ``Kxe5'', ``push the pawn'', or ``move the king''.
\textit{Spatial} markers capture deictic and positional board language, such as ``here'', ``there'', ``in front of'', ``behind'', ``toward'', ``left'', or ``right''.
\textit{Concept} markers cover tactical, strategic, and endgame vocabulary, such as ``opposition'', ``passed pawn'', ``promotion'', ``zugzwang'', ``fork'', ``check'', ``blockade'', or ``rook activity''.

Unlike the lexical categories above, specificity is a normalized text-level heuristic rather than an independent lexical category. It rewards explicit square and move notation, tactical terminology, conditional reasoning, and concrete move sequences, while penalizing generic strategic statements. The resulting score is normalized by segment length and measures textual concreteness rather than chess correctness or explanation faithfulness. Details on the specificity score are described in Appendix \ref{sec:specificity_heuristic}. 

\begin{table}[t]
\centering
\scriptsize
\setlength{\tabcolsep}{2.4pt}
\renewcommand{\arraystretch}{0.95}
\begin{tabular}{lrrrrrrr}
\toprule
Group & Seg. & Words & Coord. & Move & Spatial & Concept & Spec. \\
\midrule
Human ctrl. & 1157 & 8.1  & 0.124 & 0.125 & 0.435 & 0.136 & 0.210 \\
Human nat.  & 417  & 17.9 & 0.142 & 0.142 & 0.955 & 0.612 & 0.122 \\
LLM brief   & 6299 & 12.6 & 0.326 & 0.389 & 0.217 & 0.177 & 0.728 \\
LLM calc.   & 17895 & 10.6 & 0.292 & 0.387 & 0.174 & 0.179 & 0.415 \\
LLM teach.  & 13826 & 13.6 & 0.389 & 0.430 & 0.336 & 0.176 & 0.458 \\
\bottomrule
\end{tabular}
\caption{
Reasoning-style markers for human and LLM chess explanations.
Human ctrl. aggregates the three controlled think-aloud transcripts; Human nat. aggregates the two public commentaries.
LLM rows aggregate explanations by prompting style.
Coord., Move, Spatial, and Concept report the proportion of units containing explicit coordinates, move notation, spatial/deictic board language, or tactical/strategic vocabulary.
Spec. is the normalized text-only specificity score.
}
\label{tab:human_llm_reasoning_segments}
\end{table}

Table~\ref{tab:human_llm_reasoning_segments} compares human and LLM reasoning-style markers with the same feature extractor.
Human sources differ in how they ground reasoning: controlled think-alouds expose candidate moves and revisions in the qualitative analysis, while public commentaries are especially spatial and concept-heavy.
LLM units contain more explicit coordinates and move notation than most human sources, but substantially less deictic language than public commentary.
Prompting also changes the LLM profile: teaching has the highest coordinate, move, and spatial rates among the three LLM conditions, whereas brief explanations have the highest normalized specificity score.
Thus, fluent explanations can expose different kinds of board-grounded information rather than differing along a single specificity axis.
Controlled-transcript statistics, qualitative examples, and the exploratory ELFEN comparison appear in Appendix~\ref{sec:human_reasoning_details}.

\section{Discussion}

The main finding is not simply that explanations tend to describe the moves they accompany. That is expected in a post-hoc explanation setting. More informative are the cases in which explanation text changes model behavior without improving solution quality.

First, random explanations are not neutral. Conditioning the scorer on a plausible explanation from another puzzle can lower the rank of the correct reference move. This shows that explanation text is not merely ignored when it is irrelevant. Instead, it can actively steer token-level preferences toward an incompatible interpretation of the position. In this sense, fluent reasoning text behaves as a consequential contextual signal, even when it is not grounded in the current board state.

Second, recognizable chess motifs do not reliably indicate correct reasoning. Explanations frequently mention concepts such as opposition, promotion, king activity, or rook activity, and motif-rich explanations are sometimes easier to decode. However, this increase in recoverability does not translate into higher move accuracy. A rationale can therefore be domain-plausible, conceptually coherent, and strongly associated with a particular action while still supporting the wrong move.

These results separate three properties that are often conflated: linguistic plausibility, action consistency, and solution correctness. An explanation may use appropriate chess terminology and align with the action eventually produced, yet still fail to distinguish that action from better legal alternatives. This is especially important for interpretability, because the presence of structured domain language can create a strong impression of reasoning even when the underlying move preference remains weak or incorrect.

The masking and decoder controls reinforce this interpretation. Unmasked explanations make generated moves easy to recover largely because they expose coordinates, notation, and direct move cues. Once those cues are removed, the remaining signal is modest and decoder-dependent. Thus, recoverability should be understood as a test of whether an explanation preserves decision-specific information, not as direct evidence that the text reconstructs the model's causal reasoning process.

Taken together, the results suggest that generated explanations should be evaluated not only for whether they are coherent with an answer, but also for whether they improve discrimination among plausible alternatives. In our setting, explanations often influence model preference and expose recognizable chess concepts, but neither effect reliably tracks correctness.

\section{Conclusion}

This work evaluated whether natural-language chess explanations provide decision-relevant information beyond surface plausibility and direct move leakage. Our results show that explanation text can influence model behavior: even a random but plausible chess explanation can reduce the probability of the correct move. At the same time, explanations that contain recognizable chess motifs or are easier to decode are not reliably more accurate.

These findings show that plausible language, agreement with the generated move, and correctness are not the same thing. An explanation can sound strategically appropriate and support the move a model produced while still failing to favor the correct move over legal alternatives. After explicit move cues are removed, the remaining signal is modest and decoder-dependent. Chess therefore lets us test whether explanations genuinely affect action preferences or merely sound convincing.

\section{Limitations}

Our evaluation measures behavioral and token-level properties of explanations rather than their underlying causal generation. Move recoverability, decoder controls, and token-level scoring test whether explanations communicate action-specific information, but they do not provide direct access to the reasoning process that produced a model's decision. Likewise, the all-legal scoring experiment uses Qwen2.5-32B as a cross-model scorer for GPT-4.1-mini explanations, so it evaluates the transferability of explanation information rather than the generator's internal computation.

Several experiments are limited in scale. The decoder-control and all-legal scoring analyses are performed on 50 puzzles, while the full recoverability experiments use 200 puzzles. Decoder performance also depends on the underlying model: GPT-4.1-mini acts as a self-decoder, whereas Llama and Qwen are cross-model decoders with different chess capabilities. We therefore focus on qualitative patterns that remain consistent across decoders rather than comparing absolute recoverability rates.

Our masking strategy also involves a trade-off. Strict masking removes explicit move notation and direct references while retaining some semantic concepts such as \emph{promotion}, our audit detects residual promotion vocabulary in 56.7\% of sampled explanations. Maximal masking removes all audited leakage classes, but may also remove legitimate decision-relevant information, making it a conservative rather than perfect estimate of explanation signal.

Several analyses are exploratory. The layerwise logit-lens results provide a descriptive view of changing reference-move compatibility across layers, but should not be interpreted as a mechanistic explanation of model reasoning. Similarly, the motif analysis measures lexical mentions rather than whether chess concepts are objectively present or correctly applied, and the regression analyses identify associations rather than causal effects. Endgame categories are also unevenly represented, so categories with fewer than ten puzzles are reported descriptively only.

Finally, the human comparison is intended as a comparison of grounding styles rather than reasoning quality. The controlled think-aloud transcripts and public YouTube commentaries are small, heterogeneous, and unmatched to the LLM evaluation positions, so they should not be interpreted as human performance baselines.

\bibliography{references}

\appendix

\section{Additional Results}

\subsection{Preference-Recoverability Details}
\label{sec:preference_recoverability_details}

Tables~\ref{tab:prob_recoverability_rank} and \ref{tab:prob_recoverability_margin} report the exact counts and mean engine loss for the rank and log-probability-margin buckets visualized in Figure~\ref{fig:token_preference_recoverability}.
\begin{table}[h]
\centering
\small
\setlength{\tabcolsep}{3pt}
\begin{tabular}{lccc}
\toprule
Reference rank & $n$ & Recoverability & Mean $\Delta$ (cp) \\
\midrule
2-3 & 191 & 0.321 & 11192.7 \\
1 & 113 & 0.352 & 14443.0 \\
4-5 & 172 & 0.412 & 12272.2 \\
>5 & 112 & 0.297 & 10501.3 \\
\bottomrule
\end{tabular}
\caption{Recoverability by reference-move rank bucket (Mistral-7B token-prob scoring; brief prompt).}
\label{tab:prob_recoverability_rank}
\end{table}

\begin{table}[h]
\centering
\small
\setlength{\tabcolsep}{3pt}
\begin{tabular}{lccc}
\toprule
Margin bucket & $n$ & Recoverability & Mean $\Delta$ (cp) \\
\midrule
-5 to -2 & 146 & 0.321 & 10190.0 \\
>=-0.5 & 136 & 0.351 & 13695.5 \\
< -5 & 242 & 0.354 & 13014.5 \\
-2 to -0.5 & 64 & 0.381 & 8920.5 \\
\bottomrule
\end{tabular}
\caption{Recoverability by reference-vs-top logprob margin bucket. Margin is reference logprob minus top logprob (closer to 0 = more internal uncertainty).}
\label{tab:prob_recoverability_margin}
\end{table}

\subsection{Token-Scoring Details}
\label{sec:token_scoring_details}

Tables~\ref{tab:tokenprob_ablation} and \ref{tab:tokenprob_ablation_large_models} report the style-conditioned 7B-scale results and the Llama-3.1-70B and Qwen2.5-32B scoring-only results summarized in Figure~\ref{fig:token_preference_recoverability}.
\begin{table*}[t]
\centering
\small
\begin{tabular}{lcccccc}
\toprule
Prompt & Top-1 & Top-3 & Top-5 & Mean rank & Median rank & Mean $\Delta$ (ref-top) \\
\midrule
brief & 0.215 & 0.540 & 0.820 & 3.46 & 3.00 & -3.70 \\
calc & 0.200 & 0.535 & 0.800 & 3.52 & 3.00 & -5.34 \\
teaching & 0.175 & 0.490 & 0.805 & 3.64 & 4.00 & -5.41 \\
\bottomrule
\end{tabular}
\caption{Token-probability ablation: reference-move rank under style-conditioned scoring prompts (open-weights model). $\Delta$ is reference logprob minus top-prob logprob; more negative means the reference move is less preferred.}
\label{tab:tokenprob_ablation}
\end{table*}
\begin{table*}[t]
\centering
\scriptsize
\setlength{\tabcolsep}{4pt}
\begin{tabular}{llrrrrrr}
\toprule
Model & Prompt & Top-1 & Top-3 & Top-5 & Mean rank & Median rank & Mean $\Delta$ (ref-top) \\
\midrule
Llama-3.1-70B (4-bit) & scoring only & 0.285 & 0.575 & 0.770 & 3.35 & 3.00 & -2.14 \\
Qwen2.5-32B & scoring only & 0.245 & 0.555 & 0.780 & 3.46 & 3.00 & -2.06 \\
\bottomrule
\end{tabular}
\caption{Reference-move ranking under scoring-only prompts for the two larger open models. $\Delta$ is the reference-move log-probability minus the top-move log-probability; values closer to zero indicate a smaller preference gap.}
\label{tab:tokenprob_ablation_large_models}
\end{table*}

\subsection{Model Comparison}

Table~\ref{tab:hosted_model_summary} summarizes results by hosted model, and Table~\ref{tab:engine_clean} reports the cleaned GPT-4.1-mini move-quality analysis used in the recoverability experiments.
Table~\ref{tab:groq_prompt_summary} then gives all 18 model-prompt conditions.
These are useful for broad behavioral comparison, but the main interpretability analyses in the paper focus on open models where token-level candidate scoring is available.
\begin{table*}[t]
\centering
\small
\begin{tabular}{lccccc}
\toprule
Model & Parsed & Legal & Exact & Avg. words & Avg. specificity \\
\midrule
GPT-4.1-mini & 0.980 & 0.512 & 0.082 & 102.6 & 0.242 \\
Llama-3.1-8B-Instant & 0.740 & 0.038 & 0.007 & 147.3 & 0.239 \\
Llama-3.3-70B-Versatile & 0.478 & 0.068 & 0.005 & 124.8 & 0.162 \\
Llama-4-Scout-17B & 0.693 & 0.060 & 0.002 & 145.6 & 0.187 \\
Kimi-K2-Instruct & 0.863 & 0.487 & 0.072 & 97.1 & 0.413 \\
Qwen3-32B & 0.060 & 0.048 & 0.007 & 145.0 & 0.241 \\
\bottomrule
\end{tabular}
\caption{Hosted-model results averaged across brief, calculation, and teaching prompts on 200 puzzles. The full 18-condition table is reported in Appendix~\ref{tab:groq_prompt_summary}.}
\label{tab:hosted_model_summary}
\end{table*}
\begin{table*}[t]
\centering
\small
\begin{tabular}{lcccc}
\toprule
Prompt & Legal (95\% CI) & Exact (95\% CI) & Median loss, legal & Mean loss, all \\
\midrule
Brief & 0.535 [.470,.605] & 0.080 [.045,.120] & 471 & 1244.8 \\
Calculation & 0.415 [.350,.480] & 0.070 [.035,.110] & 459 & 1397.8 \\
Teaching & 0.585 [.515,.655] & 0.095 [.055,.140] & 463 & 1157.7 \\
\bottomrule
\end{tabular}
\caption{GPT-4.1-mini move quality on 200 puzzles. Engine loss is clipped at 2000 centipawns. ``Legal'' loss excludes illegal and unparsed outputs; ``all'' assigns them the 2000-centipawn cap. Exact match is measured over all outputs.}
\label{tab:engine_clean}
\end{table*}
\begin{table*}[t]
\centering
\small
\resizebox{\textwidth}{!}{%
\begin{tabular}{l l c c c c c c c}
\toprule
Model & Prompt & $n$ & Parsed & Legal & Exact & Avg $\Delta$ (cp) & Avg words & Avg spec. \\
\midrule
gpt-4.1-mini & brief & 200 & 0.970 & 0.535 & 0.080 & 11559.5 & 44.4 & 0.201 \\
gpt-4.1-mini & calc & 200 & 0.975 & 0.415 & 0.070 & 12364.2 & 146.1 & 0.321 \\
gpt-4.1-mini & teaching & 200 & 0.995 & 0.585 & 0.095 & 12304.3 & 117.4 & 0.205 \\
llama-3.1-8b-instant & brief & 200 & 0.960 & 0.015 & 0.005 & 66600.3 & 47.2 & 0.256 \\
llama-3.1-8b-instant & calc & 200 & 0.270 & 0.070 & 0.010 & 21781.6 & 200.7 & 0.231 \\
llama-3.1-8b-instant & teaching & 200 & 0.990 & 0.030 & 0.005 & 50084.5 & 194.1 & 0.230 \\
llama-3.3-70b-versatile & brief & 200 & 0.250 & 0.045 & 0.000 & 22378.6 & 36.8 & 0.131 \\
llama-3.3-70b-versatile & calc & 200 & 0.630 & 0.100 & 0.010 & 5283.1 & 174.9 & 0.209 \\
llama-3.3-70b-versatile & teaching & 200 & 0.555 & 0.060 & 0.005 & 16795.8 & 162.8 & 0.145 \\
meta-llama/llama-4-scout-17b-16e-instruct & brief & 200 & 0.730 & 0.100 & 0.005 & 20339.7 & 77.4 & 0.210 \\
meta-llama/llama-4-scout-17b-16e-instruct & calc & 200 & 0.500 & 0.035 & 0.000 & 28948.4 & 163.7 & 0.151 \\
meta-llama/llama-4-scout-17b-16e-instruct & teaching & 200 & 0.850 & 0.045 & 0.000 & 11368.6 & 195.7 & 0.201 \\
moonshotai/kimi-k2-instruct-0905 & brief & 200 & 0.860 & 0.505 & 0.080 & 14197.0 & 47.1 & 0.442 \\
moonshotai/kimi-k2-instruct-0905 & calc & 200 & 0.880 & 0.520 & 0.060 & 13774.0 & 125.7 & 0.477 \\
moonshotai/kimi-k2-instruct-0905 & teaching & 200 & 0.850 & 0.435 & 0.075 & 16462.0 & 118.6 & 0.319 \\
qwen/qwen3-32b & brief & 200 & 0.075 & 0.060 & 0.010 & 8571.2 & 144.6 & 0.264 \\
qwen/qwen3-32b & calc & 200 & 0.040 & 0.020 & 0.005 & 25397.5 & 138.4 & 0.214 \\
qwen/qwen3-32b & teaching & 200 & 0.065 & 0.065 & 0.005 & 7983.1 & 152.0 & 0.244 \\
\bottomrule
\end{tabular}
}
\caption{Model comparison on 200 Lichess endgame puzzles. Parsed = fraction of outputs where a move could be extracted.}
\label{tab:groq_prompt_summary}
\end{table*}

The hosted-model comparison reveals two distinct failure modes.
GPT-4.1-mini and Kimi-K2 usually emit a parseable move and reach legal-move rates near one half, whereas several open hosted models frequently produce an unparsable or illegal action despite fluent explanatory text.
Exact reference accuracy remains below 10\% in almost every model-prompt cell.
The calculation and teaching prompts substantially increase explanation length, but do not produce a consistent gain in exact accuracy or legality.
Consequently, apparent explanatory detail should not be interpreted as evidence of improved chess calculation.
The very large raw engine losses in the full condition table also reflect mate-score conversion and catastrophic legal errors; the cleaned and clipped engine analysis in Table~\ref{tab:engine_clean} is the preferred measure of move quality.

\subsection{All-Legal Scoring Visualization}

For this experiment, Qwen2.5-32B scores every legal UCI move rather than selecting among a small filtered set.
This removes candidate-set dependence and makes the rank comparisons directly interpretable.
With FEN alone, reference top-1 accuracy ranges from 0.16 to 0.18 across prompt groups.
Adding a maximally masked explanation changes these values only marginally, while unmasked text sharply raises generated-move top-1 accuracy to 0.40-0.60.
The latter increase is an expected leakage sanity check: the original move or its destination often remains recoverable from unmasked prose.
Random explanations generally lower reference ranking, demonstrating that irrelevant but plausible chess language can actively distract the scorer rather than merely add noise.

\begin{figure*}[t]
\centering
\includegraphics[width=0.7\textwidth]{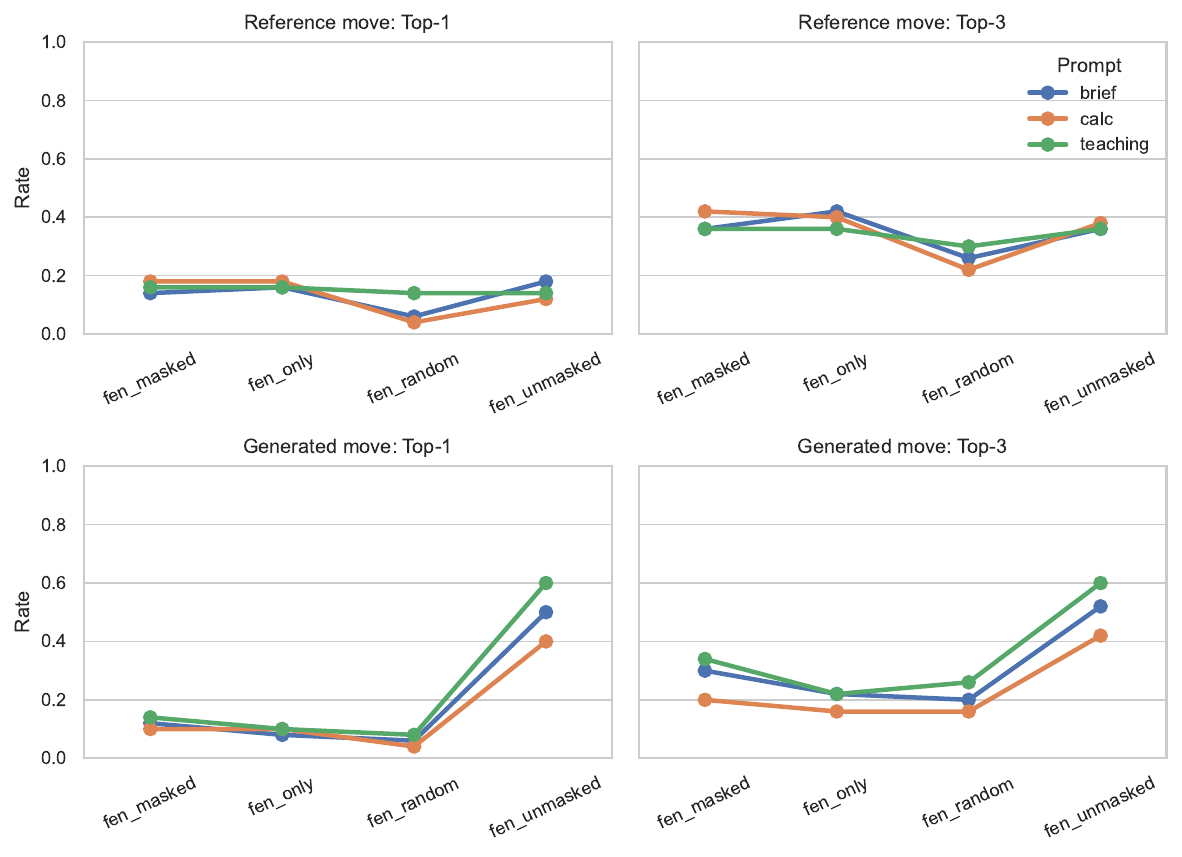}
\caption{Qwen2.5-32B all-legal-move scoring under FEN-only and explanation controls. The scorer conditions on GPT-4.1-mini explanations.}
\label{fig:qwen32b_controls}
\end{figure*}

\subsection{Counterfactual Sensitivity}

We perturb 25 puzzles by shifting a king or pawn while preserving a legal position.
The generated move changes in 96\% of variants and the top-probability move changes in 100\%, while explanation similarity remains low but non-zero.
These results suggest that generated decisions are sensitive to small board changes, but explanation decodability is less clearly coupled to those changes.

Each perturbation is deliberately local: it modifies one piece while retaining the broad material class and a legal side-to-move configuration.
This is not intended to preserve the original tactical solution; instead, it tests whether the model reacts to a nearby board state or repeats a generic narrative.
The mean text similarity of 0.203 indicates that explanations usually change lexically, but some phrasing and strategic templates persist.
The decoded move changes in 81.6\% of variants, whereas the binary recoverability match never changes because recovery is already sparse in this subset.
Thus the match indicator has a floor effect and should be read together with the decoded-action and probability-rank changes.
\begin{table}[h]
\centering
\small
\setlength{\tabcolsep}{4pt}
\begin{tabular}{lc}
\toprule
Metric & Value \\
\midrule
Mean explanation similarity & 0.203 \\
Generated move changed & 0.959 \\
Top-prob move changed & 1.000 \\
Recoverability predicted move changed & 0.816 \\
Recoverability match changed & 0.000 \\
\bottomrule
\end{tabular}
\caption{Counterfactual sensitivity summary on 25 puzzles (Mistral-7B). Values are fraction of variants where a change occurred, except explanation similarity (SequenceMatcher ratio).}
\label{tab:counterfactual_sensitivity}
\end{table}

\begin{figure}[t]
\centering
\includegraphics[width=\columnwidth]{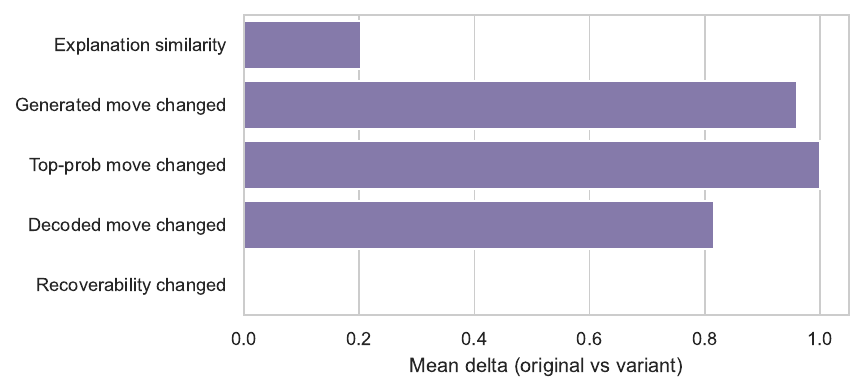}
\caption{Counterfactual sensitivity summary for 25 puzzles.}
\label{fig:counterfactual_sensitivity}
\end{figure}

\subsection{Layerwise Exploratory Details}
\label{sec:logit_lens_appendix}

For Llama-3.1-70B and Qwen2.5-32B, we apply the model output head at each layer and compute the log-probability of the complete reference-move token sequence.
Both models show increasing reference-move salience across layers, although the move generally remains low probability.
Because direct logit-lens projections can be distorted by representational changes across layers, these results are descriptive rather than causal.

The average reference-sequence log-probability rises from $-47.59$ to $-21.46$ for Llama and from $-48.26$ to $-17.64$ for Qwen.
This late-layer increase means that the reference string becomes less incompatible with the model representation as computation proceeds.
It does not mean that the model has selected the solution: the reference can improve substantially while remaining below competing legal moves.
The probe also scores the full multi-token UCI sequence, so changes may partly reflect improved prediction of move syntax rather than chess-specific planning.
\begin{table}[t]
\centering
\scriptsize
\begin{tabular}{lcccc}
\toprule
Model & Layers & Mean logprob (first) & Mean logprob (last) \\
\midrule
Llama-3.1-70B & 0-80 & -47.59 & -21.46 \\
Qwen2.5-32B & 0-64 & -48.26 & -17.64 \\
\bottomrule
\end{tabular}
\caption{Logit-lens summary (20 puzzles). Mean reference-move logprob (Lichess solution) at the first and last layer.}
\label{tab:logit_lens_summary}
\end{table}

\begin{figure}[t]
\centering
\includegraphics[width=\columnwidth]{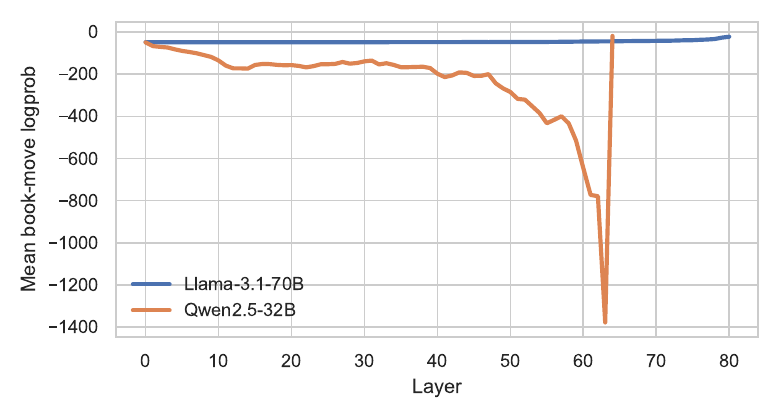}
\caption{Exploratory logit-lens reference-move logprob across layers for 20 puzzles.}
\label{fig:logit_lens_bookmove}
\end{figure}

\begin{figure}[t]
\centering
\includegraphics[width=\columnwidth]{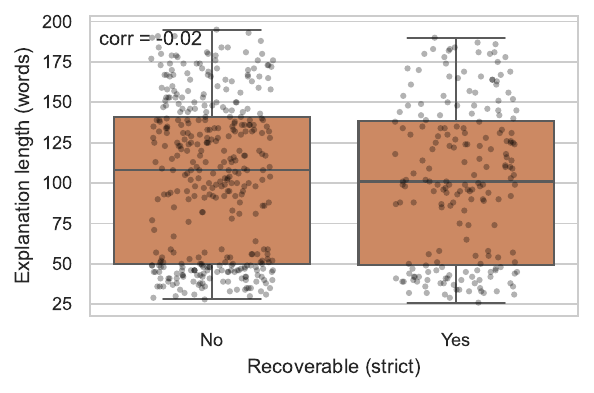}
\caption{Explanation length vs. recoverability under strict masking.}
\label{fig:length_vs_recoverability}
\end{figure}

Figure~\ref{fig:length_vs_recoverability} provides a complementary behavioral check.
Long explanations are not reliably easier to decode after move masking.
Calculation prompts generate substantially more words and more move-like lines, yet their recoverability is comparable to or below the brief condition.
Length therefore captures verbosity and surface structure, not necessarily action-specific information.

\subsection{Text-Only Specificity Heuristic}
\label{sec:specificity_heuristic}

Specificity is a composite measure of textual concreteness, not a separate lexical category and not a measure of chess correctness.
For a text segment $x$, we compute
\begin{align}
R(x) &= 2S+1.5M+1.5C+T \notag\\
     &\quad +1.5L-G, \\
\operatorname{Spec}(x) &= \frac{R(x)}{\max(1,W)},
\end{align}
where $S$ is the number of explicit square references, $M$ is the number of SAN- or UCI-like move mentions, $C$ counts conditional markers, $T$ counts tactical or endgame terms, $L$ estimates the maximum depth of an explicit move sequence, $G$ counts generic strategic phrases, and $W$ is the segment word count.
Square references match coordinates such as \texttt{e4}; move mentions include strings such as \texttt{Kc2}, \texttt{Qh7+}, and \texttt{e2e4}.
Conditional markers include terms such as ``if'', ``then'', ``after'', and ``because''.
The tactical dictionary includes terms such as ``fork'', ``opposition'', ``promotion'', ``zugzwang'', and ``passed pawn'', whereas the generic-phrase dictionary includes expressions such as ``gain an advantage'', ``improve the position'', and ``maintain control''.

The weights were fixed manually before this analysis and were not learned from correctness, recoverability, or human ratings.
Explicit squares receive the largest weight because they provide the clearest text-only board anchor.
Move mentions, conditional structure, and concrete line depth receive intermediate weight; tactical terms receive less because motif vocabulary can remain generic or be incorrectly applied.
Generic phrases receive a negative weight because they add length without necessarily identifying a position-specific action.
These choices make the score an interpretable heuristic rather than a validated linguistic or cognitive scale.

Several limitations follow.
Square and move patterns can overlap, so notation may contribute to more than one component.
The score does not test whether a referenced square, move, or motif is correct, and it does not reward deictic grounding such as ``here'', ``this side'', or ``in front'' unless explicit notation is also present.
Consequently, spoken reasoning transcribed with ASR may score lower than notation-heavy model text even when it is grounded in the visible board.

\subsection{Masking Audit}

We audit 60 explanations under four deterministic masks.
The automatic detector checks residual UCI/SAN, square coordinates, piece-square phrases, and promotion language.
Table~\ref{tab:masking_audit} shows that the maximal condition used in the all-legal scoring experiment removes all audited leakage.
The less aggressive masks preserve promotion vocabulary by design, explaining their higher audit rate.

Light masking removes the target move and obvious move labels, while strict masking additionally targets SAN/UCI-like strings.
The coordinate condition removes square references but intentionally retains broader lexical clues such as ``promotion'' or ``move the king in front of the pawn.''
Because the audit detector counts those clues, the first three conditions share a 0.567 residual flag rate.
Maximal masking removes all detector-defined classes, but this should not be read as proof that no semantic information remains: strategic descriptions can still narrow the intended action without naming a piece or square.
We therefore treat maximal masking as a conservative leakage control rather than a perfect deletion of move content.
\begin{table}[t]
\centering
\small
\begin{tabular}{lrr}
\toprule
Mask & Audited & Residual leakage \\
\midrule
Light & 60 & 0.567 \\
Strict & 60 & 0.567 \\
Coordinates & 60 & 0.567 \\
Maximal & 60 & 0.000 \\
\bottomrule
\end{tabular}
\caption{Automatic masking audit on 60 sampled explanations. The residual detector flags UCI/SAN, coordinates, piece-square phrases, and promotion language. Maximal masking removes all audited leakage classes.}
\label{tab:masking_audit}
\end{table}

\subsection{Decoder-Control Visualization}

Full-set strict-mask results for GPT-4.1-mini, Llama-3.3-70B, and Qwen3-32B appear in Table~\ref{tab:decoder_recoverability_200}.
Unparsed outputs count as failures.
Table~\ref{tab:decoder_recoverability_200_parsed} conditions on at least one extracted UCI move; the generated-over-reference ordering is unchanged.

Absolute recoverability depends strongly on the decoder.
GPT-4.1-mini recovers the generated move at top-1 in roughly 38-39\% of cases, compared with 6-10\% for the two Groq-hosted decoders.
Reference recovery is much lower for every decoder, reaching at most 9\% top-1.
This gap is the robust result: explanations communicate the model's asserted action more effectively than the puzzle solution, even when the asserted action is wrong.
The parsed-only table changes rates only modestly, so the ordering cannot be explained solely by formatting failures.
\begin{table*}[t]
\centering
\scriptsize
\resizebox{\textwidth}{!}{%
\begin{tabular}{llccccc}
\toprule
Decoder & Prompt & Parsed & Generated@1 & Generated@3 & Reference@1 & Reference@3 \\
\midrule
GPT-4.1-mini & Brief & 0.955 [0.925,0.980] & 0.392 [0.325,0.464] & 0.510 [0.438,0.577] & 0.055 [0.025,0.090] & 0.095 [0.060,0.135] \\
GPT-4.1-mini & Calculation & 0.975 [0.950,0.995] & 0.379 [0.313,0.446] & 0.462 [0.390,0.533] & 0.055 [0.025,0.085] & 0.100 [0.060,0.145] \\
GPT-4.1-mini & Teaching & 0.985 [0.965,1.000] & 0.387 [0.322,0.452] & 0.497 [0.427,0.568] & 0.090 [0.055,0.135] & 0.135 [0.090,0.185] \\
Llama-3.3-70B & Brief & 0.995 [0.985,1.000] & 0.067 [0.036,0.103] & 0.129 [0.088,0.180] & 0.010 [0.000,0.025] & 0.035 [0.010,0.065] \\
Llama-3.3-70B & Calculation & 1.000 [1.000,1.000] & 0.077 [0.041,0.118] & 0.128 [0.082,0.179] & 0.005 [0.000,0.015] & 0.030 [0.010,0.055] \\
Llama-3.3-70B & Teaching & 1.000 [1.000,1.000] & 0.075 [0.040,0.116] & 0.126 [0.080,0.176] & 0.025 [0.005,0.050] & 0.045 [0.020,0.075] \\
Qwen3-32B & Brief & 0.940 [0.905,0.970] & 0.057 [0.026,0.093] & 0.077 [0.041,0.119] & 0.010 [0.000,0.025] & 0.020 [0.005,0.040] \\
Qwen3-32B & Calculation & 0.950 [0.920,0.980] & 0.097 [0.056,0.138] & 0.159 [0.108,0.210] & 0.015 [0.000,0.035] & 0.040 [0.015,0.070] \\
Qwen3-32B & Teaching & 0.960 [0.930,0.985] & 0.060 [0.030,0.095] & 0.106 [0.060,0.146] & 0.020 [0.005,0.040] & 0.035 [0.010,0.065] \\
\bottomrule
\end{tabular}
}
\caption{Strict-mask move recoverability on 200 puzzles. Unparsed decoder outputs count as failures; brackets show puzzle-bootstrap 95\% confidence intervals.}
\label{tab:decoder_recoverability_200}
\end{table*}

\begin{table*}[t]
\centering
\scriptsize
\resizebox{\textwidth}{!}{%
\begin{tabular}{llccccc}
\toprule
Decoder & Prompt & Parsed & Generated@1 & Generated@3 & Reference@1 & Reference@3 \\
\midrule
GPT-4.1-mini & Brief & 1.000 [1.000,1.000] & 0.398 [0.330,0.466] & 0.518 [0.450,0.592] & 0.058 [0.026,0.089] & 0.099 [0.058,0.147] \\
GPT-4.1-mini & Calculation & 1.000 [1.000,1.000] & 0.379 [0.313,0.446] & 0.462 [0.390,0.533] & 0.056 [0.026,0.087] & 0.103 [0.062,0.149] \\
GPT-4.1-mini & Teaching & 1.000 [1.000,1.000] & 0.391 [0.325,0.462] & 0.503 [0.431,0.574] & 0.091 [0.056,0.137] & 0.137 [0.091,0.188] \\
Llama-3.3-70B & Brief & 1.000 [1.000,1.000] & 0.067 [0.031,0.104] & 0.130 [0.083,0.181] & 0.010 [0.000,0.025] & 0.035 [0.015,0.065] \\
Llama-3.3-70B & Calculation & 1.000 [1.000,1.000] & 0.077 [0.041,0.118] & 0.128 [0.082,0.179] & 0.005 [0.000,0.015] & 0.030 [0.010,0.055] \\
Llama-3.3-70B & Teaching & 1.000 [1.000,1.000] & 0.075 [0.040,0.116] & 0.126 [0.080,0.176] & 0.025 [0.005,0.050] & 0.045 [0.020,0.075] \\
Qwen3-32B & Brief & 1.000 [1.000,1.000] & 0.060 [0.027,0.098] & 0.082 [0.044,0.126] & 0.011 [0.000,0.027] & 0.021 [0.005,0.043] \\
Qwen3-32B & Calculation & 1.000 [1.000,1.000] & 0.103 [0.065,0.146] & 0.168 [0.119,0.222] & 0.016 [0.000,0.037] & 0.042 [0.016,0.074] \\
Qwen3-32B & Teaching & 1.000 [1.000,1.000] & 0.063 [0.031,0.099] & 0.110 [0.068,0.157] & 0.021 [0.005,0.042] & 0.036 [0.010,0.068] \\
\bottomrule
\end{tabular}
}
\caption{Parsed-only strict-mask recoverability on 200 puzzles. Rates are conditional on extracting at least one UCI move.}
\label{tab:decoder_recoverability_200_parsed}
\end{table*}

\begin{figure*}[t]
\centering
\includegraphics[width=0.78\textwidth]{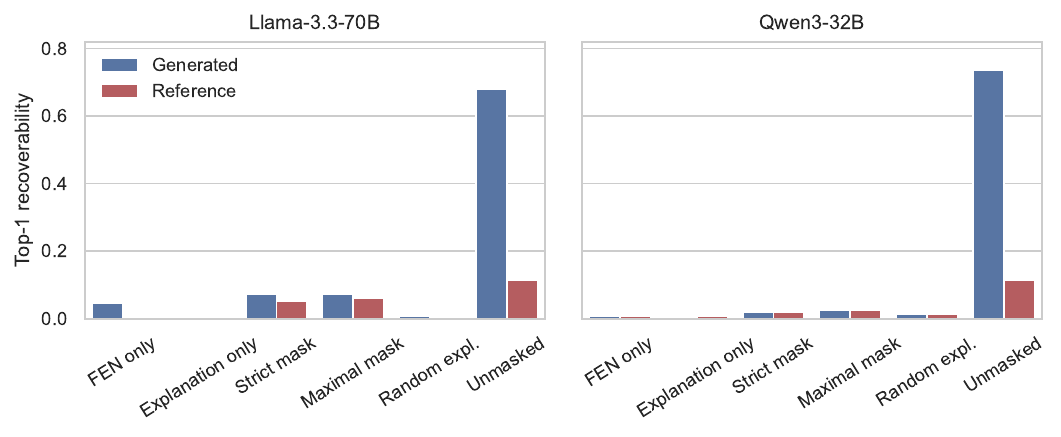}
\caption{Decoder-side top-1 recoverability controls on 50 puzzles, pooled across prompt styles. Unparsed outputs count as failures.}
\label{fig:decoder_controls_groq}
\end{figure*}

Figure~\ref{fig:decoder_controls_groq} visualizes top-1 generated- and reference-move recoverability for the two additional decoders.
The large unmasked gap is stable across models; masked conditions remain close to FEN-only and random baselines.
Table~\ref{tab:decoder_controls_groq_parsed} reports the corresponding parsed-only analysis.
Conditioning on output compliance raises absolute rates but does not turn masked explanations into reliable predictors of either target.

The unmasked condition is especially diagnostic: among parsed outputs, generated-move recovery reaches 0.779 for Llama and 0.894 for Qwen.
By comparison, strict and maximal masks remain below 0.18.
This confirms that the decoder can perform the task when the answer is lexically available; low masked recovery is therefore not simply a failure to understand UCI formatting.
Explanation-only and random-explanation conditions are near floor, indicating that useful action information arises from the interaction between board state and explanation rather than generic chess vocabulary alone.
\begin{table*}[t]
\centering
\scriptsize
\begin{tabular}{llccccc}
\toprule
Decoder & Condition & Parsed $n$ & Generated@1 & Generated@3 & Reference@1 & Reference@3 \\
\midrule
Llama-3.3-70B & FEN only & 76 & 0.092 & 0.118 & 0.000 & 0.000 \\
 & Explanation only & 30 & 0.000 & 0.100 & 0.000 & 0.033 \\
 & Strict mask & 65 & 0.169 & 0.231 & 0.123 & 0.123 \\
 & Maximal mask & 63 & 0.175 & 0.175 & 0.143 & 0.143 \\
 & Random expl. & 56 & 0.018 & 0.036 & 0.000 & 0.018 \\
 & Unmasked & 131 & 0.779 & 0.847 & 0.130 & 0.168 \\
\midrule
Qwen3-32B & FEN only & 56 & 0.018 & 0.018 & 0.018 & 0.054 \\
 & Explanation only & 18 & 0.000 & 0.056 & 0.056 & 0.056 \\
 & Strict mask & 55 & 0.055 & 0.091 & 0.055 & 0.073 \\
 & Maximal mask & 67 & 0.060 & 0.060 & 0.060 & 0.060 \\
 & Random expl. & 58 & 0.034 & 0.052 & 0.034 & 0.034 \\
 & Unmasked & 123 & 0.894 & 0.902 & 0.138 & 0.138 \\
\bottomrule
\end{tabular}
\caption{Parsed-only decoder controls on the 50-puzzle subset, pooled across prompt styles. Rates condition on at least one extracted UCI move; end-to-end rates appear in Table~\ref{tab:decoder_controls_groq}.}
\label{tab:decoder_controls_groq_parsed}
\end{table*}

\subsection{Endgame and Motif Details}

\begin{figure*}[t]
\centering
\includegraphics[width=0.32\textwidth]{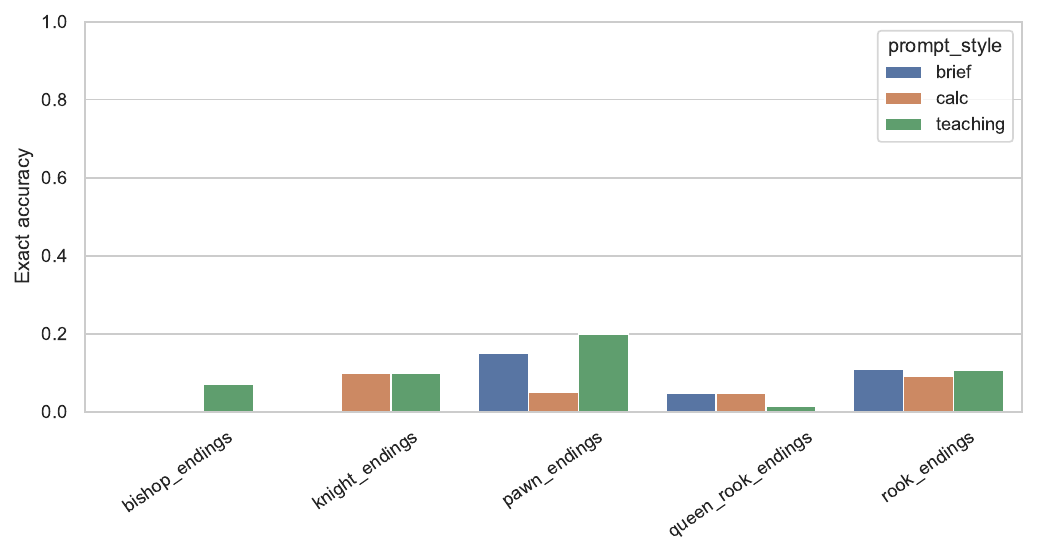}
\includegraphics[width=0.32\textwidth]{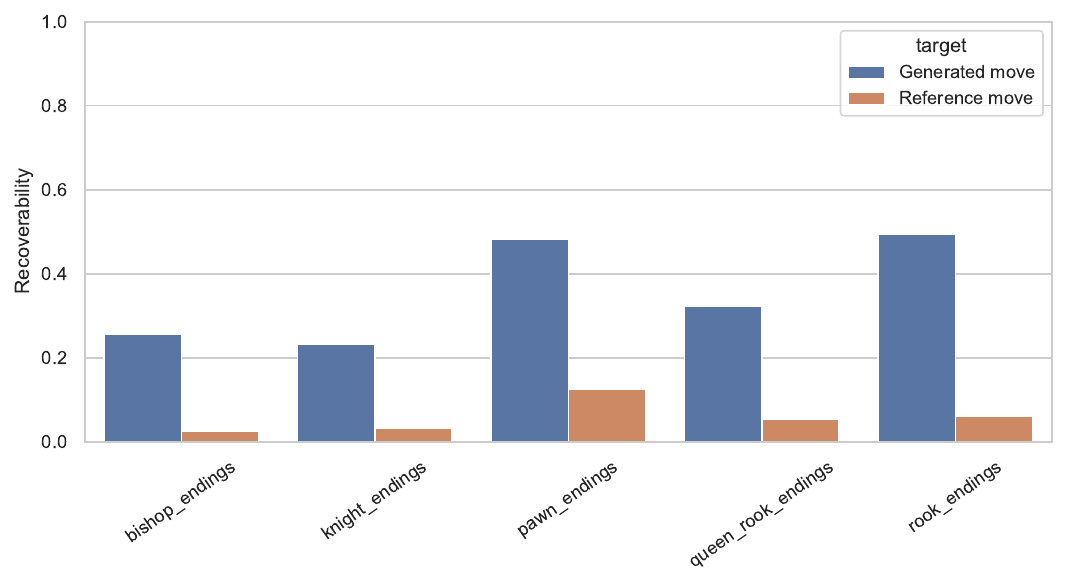}
\includegraphics[width=0.32\textwidth]{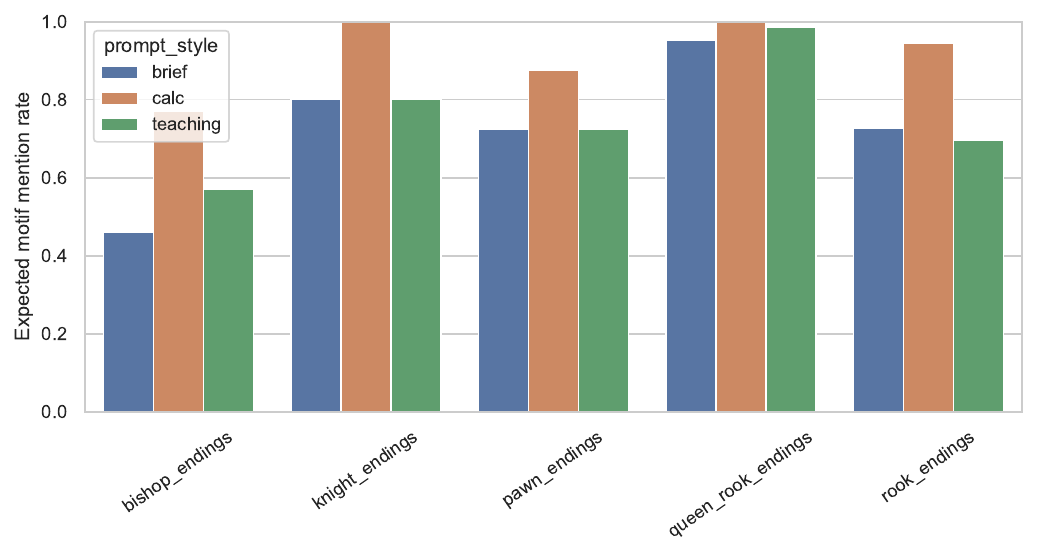}
\caption{Exact accuracy, recoverability, and expected-motif mention rates by endgame category and prompt style. Small categories should be interpreted cautiously.}
\label{fig:endgame_appendix}
\end{figure*}

\begin{figure}[t]
\centering
\includegraphics[width=\columnwidth]{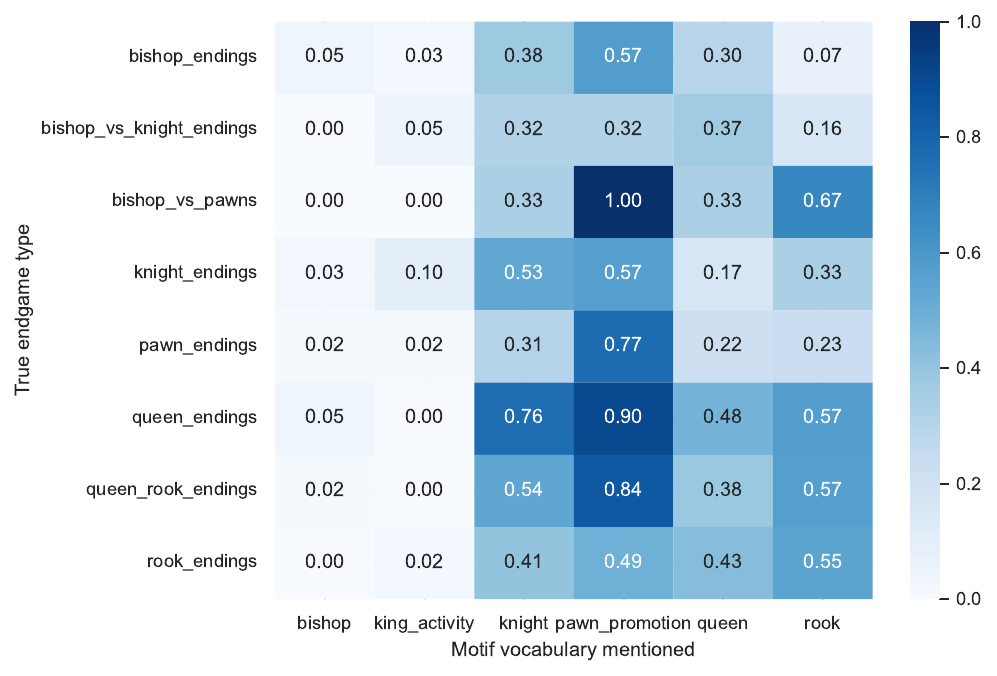}
\caption{Rows are labeled endgame types; columns are motif vocabularies mentioned in explanations.}
\label{fig:motif_confusion}
\end{figure}

Figures~\ref{fig:endgame_appendix}-\ref{fig:motif_confusion} visualize the endgame-category results.
The bar plots apply a minimum of ten puzzles per category; the confusion matrix retains all categories descriptively.
The confusion matrix shows frequent promotion and king-activity language across categories and illustrates why motif mentions alone are not evidence of correct calculation.
Table~\ref{tab:motif_effects} gives the detailed correctness and recoverability split by expected-motif mention.

Pawn endings provide the clearest separation between verbal schema recognition and exact calculation.
Mentioning an expected pawn-ending motif raises generated-move recoverability from 0.259 to 0.548, but exact accuracy rises only from 0.111 to 0.140.
For rook endings, motif mention likewise raises recoverability from 0.400 to 0.520 while exact accuracy decreases from 0.143 to 0.092.
The model can therefore produce language associated with the selected plan without selecting the reference move.
Bishop and knight cells are smaller and should be treated descriptively; their rates do not support a stable directional claim.

The confusion matrix further shows that motif vocabularies are not exclusive labels.
Promotion, king activity, checks, and captures can legitimately occur in several material classes, and generic teaching language may invoke them even when they are not decisive.
Accordingly, expected-motif mention is a lexical compatibility measure, not an annotation of tactical truth.
\begin{table}[t]
\centering
\small
\begin{tabular}{llrrr}
\toprule
Type & Expected motif & $n$ & Exact & Gen.\ rec. \\
\midrule
Pawn & absent & 27 & 0.111 & 0.259 \\
 & present & 93 & 0.140 & 0.548 \\
Rook & absent & 35 & 0.143 & 0.400 \\
 & present & 130 & 0.092 & 0.520 \\
Bishop & absent & 16 & 0.063 & 0.250 \\
 & present & 24 & 0.000 & 0.261 \\
Knight & absent & 4 & 0.000 & 0.250 \\
 & present & 26 & 0.077 & 0.231 \\
\bottomrule
\end{tabular}
\caption{Move correctness and generated-move recoverability conditional on mentioning an expected endgame motif. Counts are explanation rows, not unique puzzles.}
\label{tab:motif_effects}
\end{table}

\subsection{Exploratory Predictive Models}

We fit logistic classifiers for exact correctness, generated-move recoverability, and reference-move recoverability.
Numeric predictors are standardized; categorical variables are one-hot encoded; and confidence intervals use puzzle-clustered bootstrap resampling.
Table~\ref{tab:predictive_selected} reports selected coefficients.
Lower reference rank predicts correctness, legal lower-loss moves are easier to recover as generated actions, and whether the generated move is itself correct strongly predicts reference recovery.
Expected-motif mention has no stable conditional association with generated recovery.

The coefficient magnitudes are on the log-odds scale and should not be compared as raw percentage-point effects.
Reference-move rank has the strongest stable negative association with correctness, as expected: when the token scorer already places the solution low, the generated move is unlikely to match it.
Generated-action recovery is positively associated with legality and negatively associated with clipped engine loss, suggesting that coherent explanations are easier to decode when the selected action is at least chess-plausible.
The negative square-mention coefficient in the correctness model should not be interpreted causally.
It may reflect longer attempted calculations on difficult or incorrectly solved positions, and it illustrates why surface specificity alone is not a faithfulness guarantee.
\begin{table*}[t]
\centering
\small
\begin{tabular}{llrr}
\toprule
Outcome & Standardized predictor & Coefficient & Bootstrap 95\% interval \\
\midrule
Correct move & Reference-move rank & $-.978$ & $[-1.552,-.547]$ \\
Correct move & Square mentions & $-.616$ & $[-1.279,-.040]$ \\
Generated recovery & Legal generated move & $.403$ & $[.136,.652]$ \\
Generated recovery & Clipped engine loss & $-.371$ & $[-.715,-.169]$ \\
Reference recovery & Generated move correct & $1.109$ & $[.835,1.585]$ \\
Generated recovery & Expected motif mentioned & $.128$ & $[-.102,.405]$ \\
\bottomrule
\end{tabular}
\caption{Selected coefficients from exploratory logistic classifiers with puzzle-clustered bootstrap intervals. Models also include length, specificity, prompt, rating, material, theme, endgame type, and endgame-motif interactions. Coefficients are descriptive rather than causal.}
\label{tab:predictive_selected}
\end{table*}

\section{Exploratory Linguistic Comparison}
\label{sec:elfen_analysis}

We use ELFEN~\citep{maurer2026elfen} to compare the linguistic form of the hosted-model explanations and human transcripts.
To control text length, each model generation contributes at most one 50-word prefix, and each human transcript is divided into non-overlapping 50-word chunks.
The resulting corpus contains 2,944 LLM chunks and 336 human chunks from five transcript sources.
For each feature, we use a two-sided Mann-Whitney U test, apply Benjamini-Hochberg correction across 113 tested features, and report Cliff's delta.
At the chunk level, 102 features differ at $q<0.05$.

This chunking procedure equalizes the amount of text seen by the feature extractor, but it does not make observations independent.
Multiple chunks come from the same speaker or model generation regime, so the very small $p$-values partly reflect repeated source-level style.
Cliff's delta is therefore more informative here than significance alone: it reports how often a randomly selected LLM chunk exceeds a randomly selected human chunk on a feature.
The comparison is intended to characterize genre differences, not to infer a universal cognitive distinction between humans and models.
\begin{table}[t]
\centering
\small
\begin{tabular}{lrrrr}
\toprule
Feature & LLM & Human & $\Delta$ & $q$ \\
\midrule
Adverbs & 1.61 & 7.29 & -0.92 & $<.001$ \\
Pronouns & 3.19 & 7.22 & -0.79 & $<.001$ \\
Hedges & 1.54 & 4.30 & -0.67 & $<.001$ \\
Tokens/sentence & 36.04 & 16.43 & 0.77 & $<.001$ \\
tree depth & 6.80 & 3.58 & 0.78 & $<.001$ \\
Determiners & 6.67 & 3.66 & 0.74 & $<.001$ \\
Nouns & 9.98 & 6.07 & 0.74 & $<.001$ \\
Gunning fog & 17.35 & 7.81 & 0.83 & $<.001$ \\
\bottomrule
\end{tabular}
\caption{Exploratory ELFEN comparison of 50-word LLM explanation chunks and five human chess-reasoning transcript sources. $\Delta$ is Cliff's delta, positive when the feature is larger in LLM text. $q$ is Benjamini-Hochberg corrected.}
\label{tab:elfen_human_llm}
\end{table}

\begin{figure}[t]
\centering
\includegraphics[width=\columnwidth]{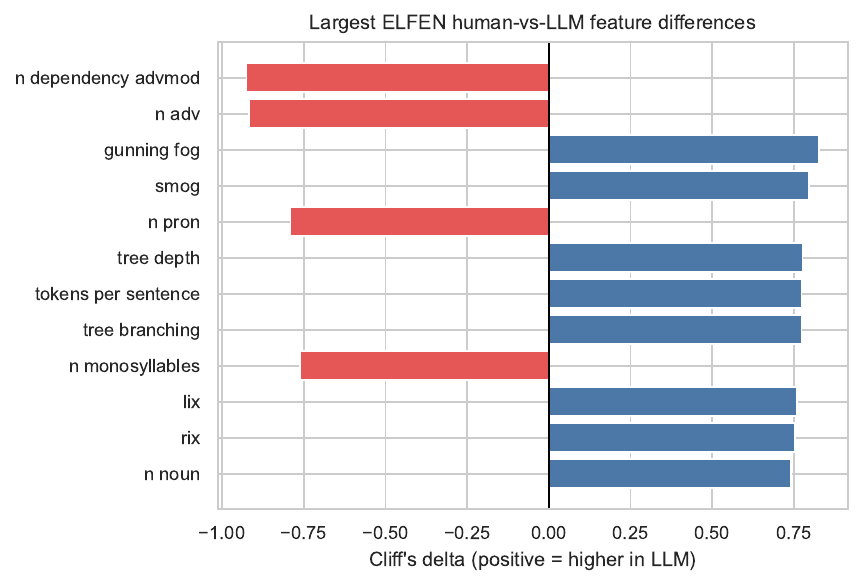}
\caption{Largest ELFEN feature differences. Positive Cliff's delta denotes a higher feature value in LLM explanations; negative values denote a higher value in human transcripts.}
\label{fig:elfen_human_llm}
\end{figure}

The largest interpretable differences distinguish spontaneous search language from polished explanation language.
Human chunks contain more adverbs, pronouns, hedges, auxiliaries, negation, and interjections.
LLM chunks contain more nouns and determiners and exhibit longer sentences, deeper dependency trees, and higher readability-complexity scores.
ELFEN's Lancaster-norm feature \citep{lynott2020lancaster} also gives LLM chunks higher average lexical visual strength ($2.52$ vs.\ $2.37$; Cliff's $\delta=0.54$), while the difference in the number of highly visual words is smaller ($\delta=0.22$).
This feature measures human ratings of words' visual associations, not whether an explanation is correctly anchored to the displayed chessboard; square and move references remain the more direct grounding measures for our task.
Figure~\ref{fig:elfen_human_llm} summarizes the largest effects.
These directions are broadly similar across the controlled and naturalistic sources, but the inferential statistics treat chunks as observations and therefore understate source-level dependence.
The comparison also confounds spoken ASR transcripts with written model output.
We consequently interpret it as exploratory stylistic evidence, not as a population-level human-LLM difference.

The transcript-level grounding analysis separates three kinds of chess language that ELFEN does not distinguish.
Coordinate/action specificity detects notation such as \texttt{Kc2} or \texttt{c4}; spatial/deictic language detects expressions such as ``here'', ``this side'', ``corner'', and ``in front''; and concept language detects tactical or strategic terms such as ``promotion'', ``capture'', ``check'', ``block'', and ``opposition.''
This distinction is important for spoken reasoning because a participant may point to a square and say ``here,'' producing a board-grounded utterance whose referent is absent from the ASR transcript.

\section{Human Transcript Excerpts}
\label{sec:human_reasoning_details}

Table~\ref{tab:human_book4_reasoning} provides the detailed controlled think-aloud summary; the segment-level comparison appears in main-text Table~\ref{tab:human_llm_reasoning_segments}.

Book-A covers four pawn endings, while Book-B and Book-C cover the first two book positions; together they form the controlled human corpus.
YouTube-J and YouTube-S are naturalistic instructional streams.
The sources differ in expertise, recording setup, segmentation, and access to a visible board, so comparisons are descriptive rather than matched estimates of human performance.
The controlled data are nevertheless useful because they expose candidate generation, opponent-response search, uncertainty, and revision before a final answer.
\begin{table*}[t]
\centering
\small
\begin{tabular}{lrrrrrrr}
\toprule
Group & $n$ & Correct & Words & Squares & Cond. & Line depth & Spec. \\
\midrule
Human pre-solution & 3 & 0.50 & 560.0 & 35.0 & 27.3 & 5.3 & 0.299 \\
Human post-solution & 3 & 1.00 & 209.7 & 21.7 & 6.7 & 7.7 & 0.606 \\
LLM brief & 4 & 0.50 & 47.5 & 1.5 & 0.0 & 1.3 & 0.152 \\
LLM calculation & 4 & 0.00 & 128.3 & 7.5 & 1.5 & 3.5 & 0.290 \\
LLM teaching & 4 & 0.00 & 126.0 & 6.8 & 0.8 & 2.3 & 0.226 \\
\bottomrule
\end{tabular}
\caption{Pilot comparison of human think-aloud reasoning and LLM explanations on book pawn endings. Human pre-solution rows are spontaneous solving attempts; post-solution rows are explanations after seeing the book line.}
\label{tab:human_book4_reasoning}
\end{table*}

Book-A is most explicit in algebraic coordinates and move sequences.
Book-B and Book-C use fewer coordinates but more deictic positional expressions, including ``here,'' ``in front,'' and ``corner.''
The YouTube sources have the highest spatial and concept-bearing segment rates, which is expected for continuous commentary accompanying a visible board.
These high rates do not imply higher solution accuracy: they measure the presence of positional or chess-concept language, not whether the referenced plan is valid.

Post-solution human explanations are also more specific than pre-solution speech because the book line supplies the move sequence.
For this reason, Table~\ref{tab:human_book4_reasoning} reports pre- and post-solution rows separately.
The pre-solution comparison is the relevant analogue of model puzzle solving; post-solution discussion instead resembles a forced-move explanation condition.

The following short excerpts illustrate the qualitative coding used in the controlled human think-aloud pilot.
We separate pre-solution excerpts from post-solution discussion because the latter is contaminated by seeing the book line.

The excerpts show two recurring contrasts.
Human speech often records abandoned candidates and uncertainty markers, while LLM text tends to compress the analysis into one confident plan.
At the same time, humans can use correct strategic vocabulary while reaching an incorrect move, so neither search-like language nor motif terminology is sufficient evidence of correctness.
\begin{table}[t]
\centering
\scriptsize
\begin{tabular}{@{}p{0.27\columnwidth}p{0.64\columnwidth}@{}}
\toprule
Source and outcome & Selected excerpt \\
\midrule
Puzzle 2, human; incorrect final answer &
``If we were to play king to d6, then we might have h5, c5, h4, c6 \ldots and then both make a queen. This might actually be a draw, so this shouldn't be right. Okay, so we can't run immediately.'' \\
\addlinespace
Puzzle 1/2, Book-B participant; concrete but wrong candidate &
``The first movement is to move the pawn from c3 into c4. \ldots I prefer to keep the king at the corner because I want to bring the white king into also corner.'' \\
\addlinespace
Puzzle 1/2, Book-C participant; schema-level plan &
``The goal of white can really probably win by promotion \ldots black wants to do is kind of just be in front of this pawn.'' \\
\addlinespace
Puzzle 2, model; \texttt{Kc7}, incorrect &
``White plays c7, pushing the pawn closer to promotion. \ldots White's king on c6 supports the pawn on c7, which is one step away from queening. \ldots Therefore, c6c7 is the winning move.'' \\
\midrule
Puzzle 4, human; unresolved &
``We can't stop this pawn from queening, I don't think so. And if a queen lands here then that also attacks our promotion square. So we might need to set it up so that our king protects that.'' \\
\addlinespace
Puzzle 4, model; \texttt{h4}, reference move &
``White should push the pawn to h4 to start advancing it towards promotion. Black's king is too far to stop the pawn \ldots This move initiates a winning plan for White.'' \\
\bottomrule
\end{tabular}
\caption{Qualitative examples from the controlled pawn-ending pilot. Human rows are think-aloud excerpts from the participants; model rows are generated explanations. Puzzle~4 is a drawing study: the model selects the reference move but incorrectly describes it as winning. Excerpts are shortened for space.}
\label{tab:human_llm_examples}
\end{table}

\end{document}